\documentclass{article}

\usepackage[utf8]{inputenc}
\usepackage[T1]{fontenc}
\usepackage{times}
\usepackage[margin=1in]{geometry}
\usepackage{hyperref}
\usepackage{url}
\usepackage{booktabs}
\usepackage{amsfonts}
\usepackage{amsmath}
\usepackage{amssymb}
\usepackage{nicefrac}
\usepackage{microtype}
\usepackage{graphicx}
\usepackage{xcolor}
\usepackage{multirow}
\usepackage{caption}
\usepackage{subcaption}
\usepackage{natbib}
\usepackage{amsthm}
\newtheorem{definition}{Definition}

\title{The Cognitive Categorical Transformer:\\ Category-Theoretic Inductive Biases for Language Modeling}

\author{
  Al Kari\\
  Manceps Inc.\\
  \texttt{research@manceps.com}
}

\date{May 2026}

\begin{document}

\maketitle

\begin{abstract}
The Cognitive Categorical Transformer (CCT) is a 306M-parameter architecture that augments a pretrained GPT-2 Small backbone with cognitively grounded components derived from category theory and several inspirations from cognitive science. Under a matched-step protocol (215{,}000 optimizer steps, matched data, matched optimizer and schedule) on WikiText-103, CCT reaches 21.27 validation perplexity, compared with 24.19 for an identically fine-tuned GPT-2 Small baseline. The architecture therefore contributes a 2.92 PPL (12\% relative) reduction beyond what in-domain fine-tuning alone provides. A retrain-from-scratch ablation that holds GT-Full simplicial message passing bypassed across the entire seven-phase activation schedule reaches 23.72 PPL, localizing 84\% of the architectural improvement (2.45 of 2.92 PPL) to GT-Full. We present the first ablation-validated evidence that simplicial message passing improves language-model perplexity at the 306M-parameter scale on WikiText-103. Published GPT-2 Large reaches 22.05 zero-shot PPL on WikiText-103 with 6.2$\times$ more parameters than GPT-2 Small; this paper treats that number as an external published reference, not as the architectural benchmark. Three negative results on consistency-style categorical priors (sheaf smoothing, adjunction round-trip, curvature regularization) and the joint structural-prior result for GT-Full and PrecisionWeightedPP together support an empirical pattern termed the \emph{structure/consistency distinction}, in which categorical priors that add new topology improve language modeling and those that enforce a consistency identity do not.
\end{abstract}

\section{Introduction}

Scaling transformer architectures \citep{vaswani2017attention} has been the dominant strategy for improving language model quality, but it imposes significant costs in parameters, compute, and energy. A complementary line of research asks whether principled structural priors, drawn from mathematics and cognitive science, can substitute topology for parameter count under a matched-step training protocol on the same data. Addressing that question rigorously requires a methodology in which the architecture being studied is compared against a fine-tuned baseline that shares its backbone initialization, its training corpus, its optimizer, its step budget, and its learning-rate schedule, so that the only difference is the architectural augmentation under test. The headline contribution of this paper is an empirical answer to that question at the 306M-parameter scale, under exactly such a matched-step comparison. Under matched-step training and matched data on WikiText-103, the Cognitive Categorical Transformer (CCT) reaches 21.27 PPL, compared with 24.19 PPL for an identically fine-tuned GPT-2 Small baseline; the difference of 2.92 PPL (12\% relative) is the architectural contribution. A retrain-from-scratch ablation in which GT-Full simplicial message passing is held bypassed across the entire seven-phase activation schedule reaches 23.72 PPL, localizing 84\% of the architectural improvement (2.45 of 2.92 PPL) to GT-Full alone.

Standard transformers treat tokens as points in a metric space, relying on dot-product attention to discover pairwise relationships. This design embodies no structural prior about language beyond positional ordering: no geometry capturing higher-order relationships, no hierarchy, and no memory beyond the context window. Category theory offers a formal language for expressing such structural relationships. Sheaves formalize the gluing of local observations into globally coherent representations. Coalgebras capture state-dependent observation processes such as memory. The Yoneda lemma characterizes objects entirely by their relationships to all other objects, providing a formal basis for self-monitoring. Simplicial complexes generalize graphs to capture higher-order relationships (triangles, tetrahedra) that pairwise attention cannot express. Each structure has precise mathematical semantics that constrain the design of neural components. Self-Determination Theory \citep{ryan2017self,sheldon2022freely} was one of several inspirations consulted during CCT's design phase, alongside predictive processing \citep{clark2013whatever,friston2010free} and the differentiable-memory tradition \citep{graves2016hybrid}; under the matched-step, static-corpus training regime evaluated here, two of seven SDT-to-architecture correspondences carry the measurable contribution (Section~\ref{sec:sdt}). The role of SDT is therefore more accurately described as a design-space constraint that helped select which structural priors to implement, not as the source of the measurable benefit.

The Cognitive Categorical Transformer (CCT) instantiates these ideas as concrete neural modules wrapping a pretrained GPT-2 backbone. A key finding is that not all categorical structures contribute equally: \emph{structural} priors (simplicial topology, precision-weighted prediction) produce substantial gains, while \emph{consistency} priors (sheaf smoothing, adjunction round-trips, curvature regularization) prove redundant or harmful. This pattern is consistent with recent sheaf-theoretic analysis of neural networks \citep{bosca2026neural}, which proves that feedforward ReLU network forward passes already achieve minimal sheaf discrepancy.

\subsection{Contributions}

\begin{enumerate}
    \item \textbf{Simplicial message passing improves language model perplexity under retrain-from-scratch ablation.} GT-Full simplicial message passing contributes 2.45 of 2.92 PPL of architectural improvement (84\%) over a fine-tuned GPT-2 Small baseline when held bypassed across all seven training phases (Table~\ref{tab:decomposition}, Figure~\ref{fig:decomposition}). The eval-only ablation share for the same component is 91.6\% (Table~\ref{tab:ablation_eval_only}), measured as learned dependence on the component's output distribution at evaluation time; the eval-only and retrain numbers measure related but distinct quantities. We present the first ablation-validated evidence that simplicial message passing improves language-model perplexity at the 306M-parameter scale on WikiText-103.

    \item \textbf{One validated structural prior, with a second under conjecture.} Of the two categorical structural priors evaluated under retrain-from-scratch ablation, GT-Full simplicial message passing is validated as the dominant architectural contributor. PrecisionWeightedPP also shows promise: its activation reduced perplexity by 1.40 PPL from Phase~6 to Phase~7 of the full CCT trajectory, but its independent contribution when GT-Full is absent is much smaller ($-0.07$ PPL in the E2 retrain ablation), suggesting that PP's benefit may be conditioned on GT-Full's structural signal. Verifying PP's independence requires a third counterfactual (CCT with GT-Full but without PP), which we leave to future work. Alongside these two structural priors, three categorical consistency priors (sheaf smoothing, adjunction round-trip loss, curvature regularization) all failed to improve language modeling, consistent with \citet{bosca2026neural}, who prove that feedforward ReLU network forward passes already minimize sheaf discrepancy on a cellular sheaf constructed from the architecture, making consistency enforcement provably redundant in that setting (Section~\ref{sec:negative_results}).

    \item \textbf{Topology substitutes for parameters under matched-step training.} CCT reaches 21.27 PPL on WikiText-103 with 306M parameters under a 215{,}000-step training budget, representing a 2.92 PPL (12\%) relative improvement over the 24.19 PPL fine-tuned GPT-2 Small baseline that consumes the identical budget without architectural augmentation (Section~\ref{sec:matched_step}). Published GPT-2 Large reaches 22.05 zero-shot PPL with 6.2$\times$ the parameter count of GPT-2 Small \citep{radford2019language}; this paper reports that number as a published external reference, not as the architectural benchmark.
\end{enumerate}

\section{Related Work}

\subsection{Category Theory in Machine Learning}

\citet{gavranovic2024categorical} established theoretical foundations for categorical deep learning at ICML, formalizing neural networks as parameterized morphisms in monoidal categories. \citet{mahadevan2026catagi} compiled the categorical framework that connects sheaves, simplicial attention, and Kan extensions to transformer architectures, and \citet{mahadevan2025categories} extends the same line of work to topos theory specifically; together these provide the mathematical infrastructure upon which CCT draws.\footnote{The simplicial attention and Kan extension constructions used here follow Mahadevan's Week 4 (``Simplicial Sets and Geometric Transformers'') and Week 5 (``Adjunctions and Kan Extensions'') of COMPSCI 692CT, Spring 2026, University of Massachusetts Amherst.} \citet{ehresmann2007memory} developed Memory Evolutive Systems using categorical colimits to model cognitive binding, anticipating several ideas in CCT's memory architecture. \citet{rosen1991life} formulated anticipatory systems using categorical closure, which connects to CCT's self-monitoring through the Yoneda lemma. Concurrent with this work, \citet{frost2026functorflow} released FunctorFlow.jl, a Julia port of Mahadevan's Python FunctorFlow package, sharing CCT's theoretical foundation in Mahadevan's categorical framework and providing an alternative, non-trained, DSL-level realization; the upstream README links the companion Lean 4 formalization at \texttt{sridharmahadevan/catagi}. The FunctorFlow GT primitive (\texttt{gt\_neighborhood\_block}, geometric message passing over simplicial structure) is the same conceptual primitive as CCT's GT-Full; the two realizations are complementary, with FunctorFlow.jl occupying the DSL/IR layer and CCT occupying the trained-model layer. The empirical contribution of this paper is the trained-model side of that division.

CCT differs from prior categorical ML work in three respects: (a) it is implemented as a complete, trainable architecture rather than a theoretical framework, (b) it grounds categorical structures in cognitive psychology rather than pure mathematics, and (c) it provides empirical evidence under retrain-from-scratch ablation that categorical inductive biases improve language modeling at the 300M parameter scale.

\subsection{Sheaf-Theoretic Foundations for Neural Networks}

\citet{bosca2026neural} prove that any feedforward ReLU network's forward pass is the unique harmonic extension of boundary data on a cellular sheaf constructed from the network's architecture. This result has a direct consequence for categorical architecture design: the forward pass already minimizes sheaf discrepancy (the Dirichlet energy on the sheaf Laplacian), so any additional sheaf consistency loss is mathematically redundant and can only conflict with the primary training objective.

\subsection{Geometric and Topological Deep Learning}

\citet{bodnar2022weisfeiler} introduced simplicial message passing networks, demonstrating that higher-order interactions capture structural information inaccessible to standard graph neural networks. \citet{hajij2025topological} provided a comprehensive framework for topological deep learning. These works operate primarily on molecular and social graphs with fixed, input-given topology. CCT adapts simplicial message passing to \emph{sequence} modeling through three modifications: (a) topology is constructed dynamically from learned coordinate projections, (b) the graph is rebuilt at every layer from the evolving representation, and (c) readout aggregates back to the sequence dimension via gated fusion with the residual stream.

\subsection{Cognitive Architectures and Bottleneck Mechanisms}

Predictive processing \citep{clark2013whatever,friston2010free} models the brain as a hierarchical prediction engine that minimizes precision-weighted prediction errors. CCT implements this as a concrete module with learned per-feature precision networks. Classical cognitive architectures (ACT-R, \citealp{anderson2007human}; SOAR, \citealp{laird2012soar}; CLARION, \citealp{sun2002duality}) implement cognitive theories at the symbolic level, whereas CCT implements cognitive principles as differentiable neural modules trained end-to-end. \citet{goyal2022inductive} advocated for cognitively-motivated inductive biases in deep learning. CCT provides a concrete instantiation of this program grounded in category theory.

\subsection{Self-Determination Theory in AI}

Self-Determination Theory \citep{ryan2017self} was one of several inspirations that informed CCT's design phase. \citet{sheldon2022freely} extends SDT with self-concordance theory and the organismic integration continuum, and \citet{sheldon2025sapient} extends SDT to artificial sapience through the Goal Breakthrough Model, which specifies how Default Mode, Salience, and Cognitive Control brain networks interact during self-determined goal selection. The Goal Breakthrough Model argues that several SDT constructs require interactive, goal-directed contexts to become computationally meaningful and would not be expected to produce measurable benefit under static-corpus training. This paper reports the empirical asymmetry honestly: two of seven SDT-to-architecture correspondences carry the measurable contribution under matched-step training on WikiText-103 (Section~\ref{sec:sdt}); the remaining five mappings are reported as unverified rather than as architectural successes. Whether they become productive under interactive training is left as an open question.

\subsection{Differentiable Memory}

\citet{graves2016hybrid} introduced the Differentiable Neural Computer (DNC). CCT's HierarchicalMemory extends the DNC with a three-tier hierarchy (buffer, working, episodic) and a batched implementation that eliminates the sequential bottleneck.

\subsection{Parameter-Efficient Methods}

LoRA \citep{hu2022lora} and adapter methods \citep{houlsby2019parameter} add lightweight modules to pretrained transformers. CCT shares the augmentation paradigm but differs in purpose: PEFT methods aim for task adaptation with minimal parameters, while CCT introduces new inductive biases intended to improve core language modeling capability. CCT's 182M added parameters reflect structural enrichment rather than parameter-efficient adaptation.

\subsection{Language Model Baselines}

Perplexity comparisons use published WikiText-103 zero-shot results: GPT-2 Small (124M, 37.50; \citealp{radford2019language}), GPT-2 Medium (355M, 26.37), GPT-2 Large (774M, 22.05), GPT-2 XL (1.5B, 17.48), and Transformer-XL (257M, 24.00; \citealp{dai2019transformer}). Downstream comparisons draw on Pythia \citep{biderman2023pythia}, OPT \citep{zhang2022opt}, and BLiMP baselines \citep{warstadt2020blimp}.

\subsection{Matched-Step Architectural Comparison Methodology}

Recent suites such as Pythia \citep{biderman2023pythia} and OPT \citep{zhang2022opt}, building on the compute-optimal training methodology of \citet{hoffmann2022chinchilla}, have demonstrated the value of controlling compute and data when comparing model families, by holding training tokens and optimizer configuration fixed across scales and architectures. The matched-step comparison reported in this paper applies the same methodological commitment at the level of a single architectural augmentation: the 24.19 PPL fine-tuned GPT-2 Small baseline (E1) and the 21.27 PPL full CCT (RC2 headline) share the GPT-2 Small initialization, the WikiText-103 training corpus, the optimizer, the batch size, the BPTT length, the AMP regime, the learning-rate schedule, and the 215{,}000-step budget; the only difference is which CCT components are wired in. The 23.72 PPL retrain-from-scratch ablation (E2) preserves all of those controls while holding GT-Full bypassed throughout the seven-phase activation schedule, making the resulting ablation delta an architectural quantity rather than a learned-dependence quantity. The eval-only ablation, which loads a fully trained CCT checkpoint and bypasses a component at forward time only, measures a related but distinct quantity (Section~\ref{sec:eval_only_carry}). Maintaining both measurements explicitly is a methodological commitment that this paper inherits from the matched-step spirit of the Pythia and OPT studies.

\section{Theoretical Framework}
\label{sec:theoretical_framework}

\subsection{Category Theory as Architectural Language}

Following \citet{mahadevan2026catagi,mahadevan2025categories} and \citet{gavranovic2024categorical}, category theory is adopted as the mathematical language for specifying the architecture.

\textbf{Design Principle (Functorial Composition).} Each CCT component is treated as a functor $F_i : \mathbf{Cog} \to \mathbf{Vect}$, where $\mathbf{Cog}$ has cognitive states as objects and cognitive transformations as morphisms, and $\mathbf{Vect}$ is the category of finite-dimensional vector spaces. The functorial constraint $F(g \circ f) = F(g) \circ F(f)$ ensures that composed components preserve compositional structure. The full CCT architecture is the composition $F_n \circ \cdots \circ F_1$.

\subsection{Sheaf-Theoretic Perception}

\begin{definition}[Perceptual Sheaf]
A perceptual sheaf on the simplicial complex $K$ of token neighborhoods assigns to each simplex $\sigma$ a vector space $\mathcal{F}(\sigma)$ together with restriction maps $\rho_{\sigma,\tau} : \mathcal{F}(\sigma) \to \mathcal{F}(\tau)$ for $\tau$ a face of $\sigma$, satisfying locality and gluing axioms. GT-Full implements this: the $k$-NN graph defines $K$, edge MLPs compute 1-simplex representations, triangle aggregation computes 2-simplex representations, and readout performs gluing.
\end{definition}

\subsection{Coalgebraic Memory}

\begin{definition}[Memory Coalgebra]
A memory system is a coalgebra for the functor $G(X) = \mathrm{Observation} \times X^{\mathrm{Action}}$, where $X$ is the set of memory states, $\mathrm{Observation}$ maps a state to what can be read, and $X^{\mathrm{Action}}$ maps a state and action to a successor state. HierarchicalMemory implements a graded version with three levels: buffer, working, and episodic.
\end{definition}

\subsection{Yoneda Self-Monitoring}

\begin{definition}[Yoneda Self-Knowledge]
The Yoneda lemma states that an object $X$ in a category $\mathcal{C}$ is completely determined by the functor $\mathrm{Hom}(X, -) : \mathcal{C} \to \mathbf{Set}$. YonedaSelfModel approximates this by maintaining $K$ probe embeddings and predicting how the model's hidden state maps to each probe. The competence signal is the accuracy of these self-predictions, measured via KL divergence.
\end{definition}

\subsection{Precision-Weighted Predictive Processing}
\label{sec:pp_definition}

\begin{definition}[Precision-Weighted Prediction]
For adjacent layers $L_i$ (higher) and $L_{i-1}$ (lower), the module computes: prediction $\hat{h}_{i-1} = f_\theta(h_i)$, error $e = h_{i-1} - \hat{h}_{i-1}$, precision $\pi = \mathrm{softplus}(g_\phi(h_i))$, and weighted error $w = \pi \odot e$. The loss
\begin{equation}
    \mathcal{L}_{\mathrm{PP}} = \tfrac{1}{2}\,\pi \odot e^2 \;-\; \tfrac{1}{2}\,\log \pi
    \label{eq:pp_loss}
\end{equation}
prevents precision collapse via the log-precision term. The architectural realization is described in Section~\ref{sec:architecture}; this paper uses Equation~\eqref{eq:pp_loss} as the unique source of the PP loss definition.
\end{definition}

\subsection{Self-Determination Theory as Architectural Specification}

Self-Determination Theory \citep{ryan2017self,sheldon2022freely} supplies the cognitive-functional specification that category theory alone does not: \emph{which} functions an architecture should implement. CCT instantiates seven correspondences between SDT constructs and categorical structures, mapping each to a concrete neural module: Grand Hierarchy to simplicial filtration (GT-Full), Symbolic Self to the Yoneda embedding (YonedaSelfModel), the autonomy / competence / relatedness needs to derived signals (BasicNeedsDashboard), Self-Concordance to coherence modulation (SelfConcordanceGate), Organismic Integration to a graded coalgebra (HierarchicalMemory), TOTE to coalgebraic unfolding (memory consolidation), and Downward Causation to top-down precision-weighted prediction (PrecisionWeightedPP). These correspondences are not post-hoc labels: each SDT construct motivated a specific architectural decision. The full mapping table is given in Appendix~A. Of the seven mappings, two carry the measurable contribution under the matched-step, static-corpus training regime evaluated here; the asymmetry is discussed honestly in Section~\ref{sec:sdt}. \citet{sheldon2025sapient} provides independent theoretical support through the Goal Breakthrough Model, which specifies how DMN, SN, and CCN brain networks interact during self-determined goal selection; the Goal Breakthrough Model's emphasis on multi-level relational processing aligns with the Grand Hierarchy mapping to GT-Full's simplicial message passing.

\section{Architecture}
\label{sec:architecture}

\subsection{Overview}

\begin{figure}[t]
\centering
\includegraphics[width=0.85\textwidth]{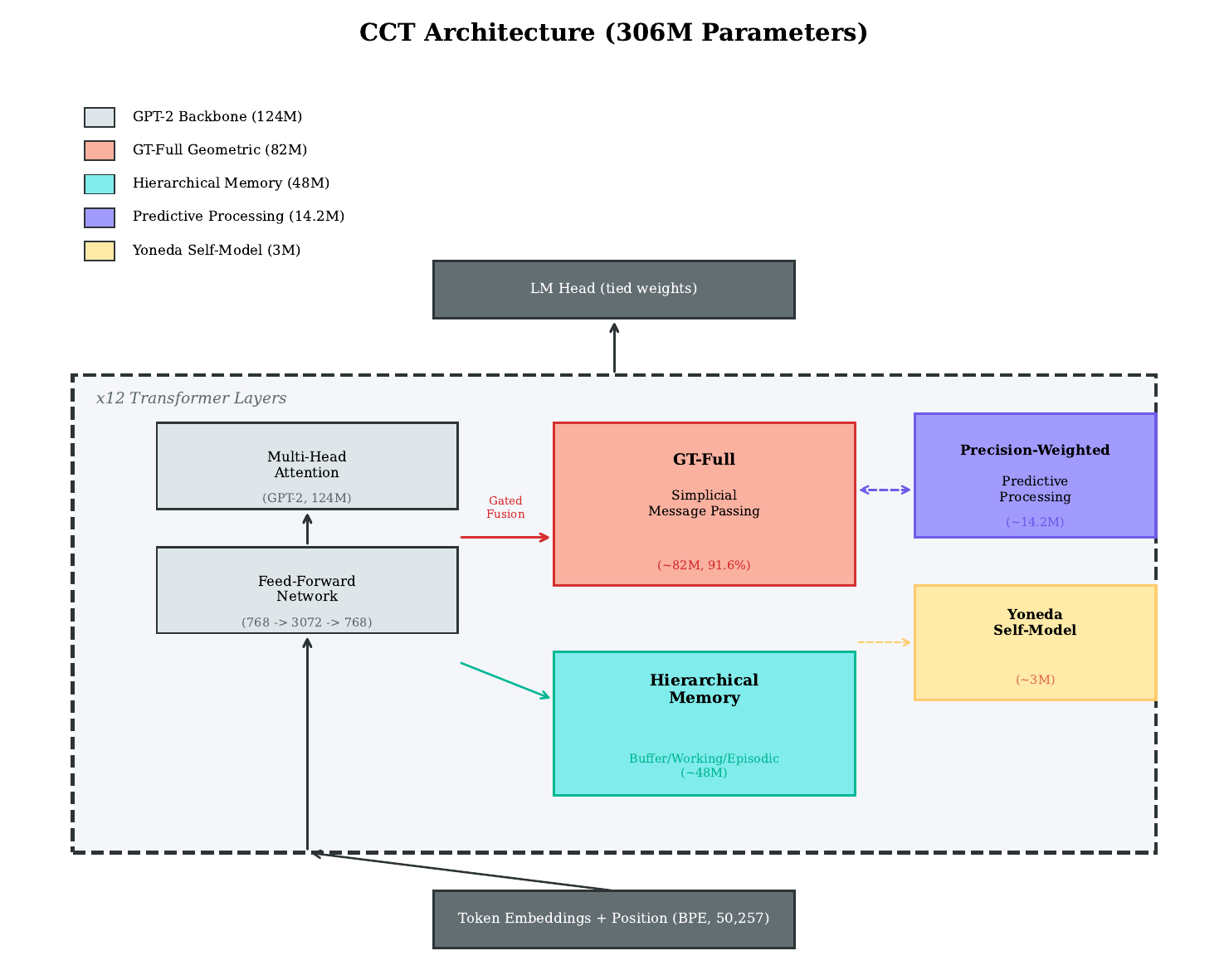}
\caption{CCT Architecture (306M parameters). The GPT-2 Small backbone (gray, 124M) is augmented with per-layer cognitive components, GT-Full simplicial message passing (coral, $\sim$82M) gated into the residual stream and a three-tier Hierarchical Memory (teal, $\sim$48M), plus model-level Precision-Weighted Predictive Processing (purple, $\sim$14.2M) with cross-layer feedback and a Yoneda Self-Model (yellow, $\sim$3M) that monitors hidden states. Figure inherited from prior work; the 91.6\% on-figure annotation reports the eval-only finding superseded by this paper's retrain ablation, and the architectural share under retrain is 84\% (Section~\ref{sec:decomposition}, Table~\ref{tab:decomposition}).}
\label{fig:architecture}
\end{figure}

CCT is a 306M-parameter model comprising a 124M-parameter GPT-2 backbone and 182M parameters of cognitive components distributed across five parameter-bearing top-level modules. The architecture is depicted in Figure~\ref{fig:architecture}; the parameter budget is enumerated in Table~\ref{tab:params}. The seven SDT-to-architecture correspondences listed in Appendix~A enumerate a larger set of named auxiliary submodules (BasicNeedsDashboard, SelfConcordanceGate, narrative GRU within YonedaSelfModel, and so forth), but the five top-level parameter-bearing modules in Table~\ref{tab:params} are what the empirical decomposition in Section~\ref{sec:results} ablates against.

\begin{table}[t]
\centering
\caption{Parameter Budget}
\label{tab:params}
\small
\begin{tabular}{lrrp{4cm}}
\toprule
\textbf{Component} & \textbf{Params} & \textbf{\%} & \textbf{Origin} \\
\midrule
GPT-2 Backbone & 124M & 40.5 & Pretrained \citep{radford2019language} \\
GT-Full ($\times$12 layers) & $\sim$82M & 26.8 & \citet{bodnar2022weisfeiler,mahadevan2026catagi} \\
HierarchicalMemory ($\times$12) & $\sim$48M & 15.7 & SDT-motivated \\
PrecisionWeightedPP & $\sim$14.2M & 4.6 & \citet{clark2013whatever,friston2010free} \\
CausalAttention ($\times$12) & $\sim$9M & 2.9 & \citet{mahadevan2026catagi} \\
YonedaSelfModel & $\sim$3M & 1.0 & Yoneda lemma \\
Other & $\sim$24M & 7.9 & Shared / overhead \\
\midrule
\textbf{Total} & \textbf{$\sim$306M} & \textbf{100} & \\
\bottomrule
\end{tabular}
\end{table}

\subsection{GPT-2 Backbone}

The backbone consists of token embedding ($50{,}257 \times 768$, BPE via tiktoken), position encoding ($1{,}024 \times 768$, learned), 12 transformer layers (LayerNorm, 12-head attention with $d_k = 64$, LayerNorm, FFN $768 \to 3{,}072 \to 768$ with GELU), and a tied output projection. All backbone weights are initialized from GPT-2 Small and trained at 50$\times$ lower learning rate than cognitive components ($5 \times 10^{-7}$ vs $1 \times 10^{-5}$); the token and position embeddings train at a further-reduced $2 \times 10^{-7}$. These three rates are used identically by RC2 Phases~2 through~7 and by the RC3 E1 and E2 phase configs.

\subsection{GT-Full: Simplicial Geometric Processing}
\label{sec:gtfull}

GT-Full is the architecture's most empirically significant component (Section~\ref{sec:ablation}). It implements simplicial message passing over dynamically constructed graphs in a learned coordinate space.

\textbf{Per-layer architecture:}
\begin{enumerate}
    \item \textbf{Coordinate projection:} Linear$(768, 384)$. Output clamped to $[-20, 20]$ for numerical stability.
    \item \textbf{$k$-NN graph construction:} $k = 6$ nearest neighbors by Euclidean distance, producing $T \times k$ edges per batch element.
    \item \textbf{Simplicial lifting:} Triangles (2-simplices) identified via GPU-accelerated sparse matrix multiplication. Given adjacency matrix $A$, $A^2$ is computed and element-wise AND with $A$ yields triangle edges. Maximum 1{,}024 triangles per pass ($\sim$2{,}500$\times$ speedup over $O(n^3)$).
    \item \textbf{Edge message passing:} Concatenated source/target coordinates processed through a 2-layer MLP ($768 \to 384 \to 384$, ReLU), aggregated per node via mean pooling.
    \item \textbf{Triangle message passing:} Three-vertex features processed through a 2-layer MLP, aggregated per node.
    \item \textbf{Gated fusion:} $h' = h + \sigma(\mathrm{gate\_bias} + \mathrm{gate\_proj}(h)) \cdot \mathrm{readout}(\mathrm{geo\_msg})$. Gate bias initialized to $-5.0$ ($\sigma(-5) \approx 0.007$); readout weights initialized to zero.
    \item \textbf{Curvature estimation:} Approximate Ollivier-Ricci curvature (diagnostic only; regularization weight $0.0$).
\end{enumerate}

\subsection{HierarchicalMemory}

Three-tier DNC-inspired \citep{graves2016hybrid} external memory: buffer (64 slots), working (32 slots), episodic (128 slots). All operations in float32 under AMP. The batched implementation computes all $T$ reads from a single memory snapshot simultaneously, recovering an approximately 250$\times$ speedup over the sequential DNC formulation.

\subsection{PrecisionWeightedPP}

For each adjacent layer pair ($L_i$ higher, $L_{i-1}$ lower), the module computes a prediction head (2-layer MLP, $768 \to 384 \to 768$), a prediction error (element-wise difference), a precision network (Linear $+$ Softplus, initialized to output $1.0$), the precision-weighted error, FiLM modulation ($\gamma$, $\beta$), and a precision-gated blend with the residual stream. Computed in float32 under AMP. The loss formulation is defined in Section~\ref{sec:pp_definition}, Equation~\eqref{eq:pp_loss}; it is not repeated here. PrecisionWeightedPP's measured marginal contribution differs sharply between the RC2 run, in which GT-Full is active, and the RC3 E2 ablation, in which it is not; the conditional-PP-on-GT-Full hypothesis arising from this contrast is analyzed in Section~\ref{sec:pp_conditional}.

\subsection{YonedaSelfModel}

Maintains 8 probe embeddings approximating the Yoneda functor. A narrative GRU (hidden dimension 128) tracks temporal evolution. The self-model auxiliary loss is computed on detached hidden states so its gradients do not destabilize the residual stream.

\subsection{Passthrough Initialization}

All cognitive components employ passthrough initialization (Table~\ref{tab:passthrough}), ensuring no regression when a new component is enabled.

\begin{table}[t]
\centering
\caption{Passthrough Initialization Scheme}
\label{tab:passthrough}
\small
\begin{tabular}{lll}
\toprule
\textbf{Component} & \textbf{Initialization} & \textbf{Effect} \\
\midrule
GT-Full gate & gate\_proj.bias $= -5.0$ & $\sigma(-5) \approx 0.007$ \\
GT-Full readout & weights $= 0$, bias $= 0$ & Zero output \\
Memory output & weights $= 0$ & Zero contribution \\
PP precision & output $= 1.0$ & Unit precision \\
\bottomrule
\end{tabular}
\end{table}

\subsection{Progressive Activation Protocol}
\label{sec:progressive_activation}

CCT components are introduced sequentially across seven phases totaling 215{,}000 optimizer steps. Each phase resumes from the best validation checkpoint of the previous phase, unfreezes one additional cognitive component, and continues training under the three-tier learning rate scheme described above. Phase~0 (or ``Baseline'') trains GPT-2 alone (all CCT side-paths bypassed via a master flag on each block); subsequent phases activate CausalAttention, GT-Full, HierarchicalMemory~+~YonedaSelfModel, the full TopDown stack, an extended-training phase at fixed component set, and finally PrecisionWeightedPP. The phase boundaries (20K / 10K / 15K / 20K / 30K / 100K / 20K steps) and the resume-from-best-checkpoint chain are identical between the RC2 experiments and the RC3 E2 retrain-from-scratch ablation, so the protocol is the shared substrate for two different ablation styles: (i) eval-only ablation, which removes a component at evaluation time from a fully trained checkpoint to measure the model's learned dependence on it, and (ii) retrain-from-scratch ablation, which re-runs the entire seven-phase chain with the component held bypassed throughout to measure the counterfactual architectural contribution. Section~\ref{sec:results} reports both.

Passthrough initialization is what makes progressive activation safe: at the moment a component is unfrozen, its output is multiplicatively close to zero, so the model starts each new phase at the previous phase's validation perplexity rather than incurring a re-adaptation penalty. The protocol is also what enables the RC3 E2 ablation to be a meaningful counterfactual: holding GT-Full bypassed across all seven phases produces a model that has been allowed the full 215K-step compute budget to compensate, isolating GT-Full's architectural contribution from any confound with training time.

\section{Experimental Setup}
\label{sec:setup}

This section describes the experimental setup used in this revision (RC3). The RC2 training configuration is carried forward without change so that all matched-step comparisons are exact, with two additions: a fine-tuned GPT-2 Small baseline (E1) and a retrain-from-scratch ablation in which GT-Full simplicial message passing is disabled across the entire seven-phase schedule (E2). Both runs use the same data, optimizer, batch size, sequence length, mixed-precision regime, learning-rate schedule, and step budget as the RC2 headline run.

\subsection{Data}
\label{sec:data}

All experiments use WikiText-103 \citep{merity2017pointer}: approximately 103M tokens of curated Wikipedia text, BPE-tokenized with the GPT-2 vocabulary (50{,}257 tokens). The corpus is split into 1{,}801{,}350 training sentences, 3{,}760 validation sentences, and 4{,}358 test sentences. Perplexity is reported on the validation split. E1 and E2 use the identical BPE-tokenized WT-103 corpus and the same shuffling seed as the RC2 run, so any difference in held-out perplexity is attributable to architecture or training trajectory rather than data preparation.

\subsection{Training Configuration}
\label{sec:training_config}

Hyperparameters are summarized in Table~\ref{tab:training_rc3}. The three-tier learning-rate structure (cognitive components at $1 \times 10^{-5}$, the pretrained GPT-2 backbone at $5 \times 10^{-7}$, and the embedding matrix at $2 \times 10^{-7}$) is the same one selected in the RC2 hyperparameter sweep and is retained verbatim. For E1, which contains no cognitive components, only the backbone and embedding learning rates are in force; this matches the RC2 baseline phase configuration and was selected so that the comparison to E2 and the RC2 full CCT run is one of architecture under matched-step, not one of optimizer or schedule.

\begin{table}[t]
\centering
\caption{Training Hyperparameters (Common to E1, E2, and the RC2 Full CCT Run). \textsuperscript{$\dagger$}Linear warmup steps differ across runs: E1 uses 500 steps; E2 phases 0, 1, 2, 3, and 6 use 1{,}000 steps; phase 4 uses 2{,}000 steps; phase 5 uses 3{,}000 steps. The warmup count is a per-phase property of the cosine schedule; all other schedule parameters (peak LRs, $\eta_{\min}$, total step budget, cosine shape) are matched across E1, E2, and RC2.}
\label{tab:training_rc3}
\small
\begin{tabular}{ll}
\toprule
\textbf{Parameter} & \textbf{Value} \\
\midrule
Optimizer & AdamW (weight\_decay $= 0.01$) \\
LR (CCT components) & $1 \times 10^{-5}$ \\
LR (pretrained backbone) & $5 \times 10^{-7}$ \\
LR (embeddings) & $2 \times 10^{-7}$ \\
Gradient clipping (max norm) & 0.5 \\
Batch size & 4 \\
Sequence length (BPTT) & 256 \\
Precision & Mixed (AMP FP16; float32 for memory, PP) \\
Scheduler & CosineAnnealingLR ($\eta_{\min} = 10^{-6}$) \\
LR warmup (linear) & E1: 500 steps; E2: per-phase (see footnote\textsuperscript{$\dagger$}) \\
Total training steps & 215{,}000 \\
Random seed & 42 (single seed) \\
\bottomrule
\end{tabular}
\end{table}

\subsection{Progressive Activation Schedule}
\label{sec:progressive}

The full CCT (RC2) and the E2 retrain ablation share the seven-phase progressive activation schedule introduced in RC2. Components are activated sequentially, with each phase resuming from the best validation checkpoint of the previous phase. The schedule for the RC2 run is given in Table~\ref{tab:phases_rc2} for reference; the E2 modification, in which GT-Full is disabled in every phase and the slot is repurposed to extend CausalAttention training, is given in Table~\ref{tab:phases_e2}.

\begin{table}[t]
\centering
\caption{RC2 Progressive Activation Schedule (Full CCT)}
\label{tab:phases_rc2}
\small
\begin{tabular}{llrl}
\toprule
\textbf{Phase} & \textbf{Components Activated} & \textbf{Steps} & \textbf{Resume From} \\
\midrule
Baseline & GPT-2 only & 20{,}000 & GPT-2 pretrained \\
Phase 2 & $+$ CausalAttention & 10{,}000 & Baseline best \\
Phase 3 & $+$ GT-Full & 15{,}000 & Phase 2 best \\
Phase 4 & $+$ Memory $+$ SelfModel & 20{,}000 & Phase 3 best \\
Phase 5 & $+$ TopDown (full stack) & 30{,}000 & Phase 4 best \\
Phase 6 & Extended training (cosine LR continued) & 100{,}000 & Phase 5 best \\
Phase 7 & $+$ PrecisionWeightedPP & 20{,}000 & Phase 6 best \\
\midrule
\textbf{Total} & & \textbf{215{,}000} & \\
\bottomrule
\end{tabular}
\end{table}

\begin{table}[t]
\centering
\caption{E2 Retrain-from-Scratch Schedule (GT-Full Disabled Throughout)}
\label{tab:phases_e2}
\small
\begin{tabular}{llrl}
\toprule
\textbf{Phase} & \textbf{Components Activated (GT-Full off)} & \textbf{Steps} & \textbf{Resume From} \\
\midrule
0 & GPT-2 only (master bypass) & 20{,}000 & GPT-2 pretrained \\
1 & $+$ CausalAttention & 10{,}000 & Phase 0 best \\
2 & CausalAttention extended (GT-Full slot diverted) & 15{,}000 & Phase 1 best \\
3 & $+$ HierarchicalMemory $+$ YonedaSelfModel & 20{,}000 & Phase 2 best \\
4 & $+$ TopDown (full stack except GT-Full) & 30{,}000 & Phase 3 best \\
5 & Extended training (cosine LR continued) & 100{,}000 & Phase 4 best \\
6 & $+$ PrecisionWeightedPP & 20{,}000 & Phase 5 best \\
\midrule
\textbf{Total} & & \textbf{215{,}000} & \\
\bottomrule
\end{tabular}
\end{table}

The E2 schedule matches the RC2 total step count (215{,}000) exactly, with the 15{,}000 steps that would have activated GT-Full in RC2 reassigned to extended CausalAttention training. Every other component (CausalAttention, HierarchicalMemory, YonedaSelfModel, TopDown, PrecisionWeightedPP) is wired in its RC2 schedule position with identical passthrough initialization. The E2 ablation is therefore a counterfactual answer to the question, ``if CCT had been built without GT-Full from the start, how much worse would it be?'' This is a strictly stronger ablation than the eval-only ablation reported in RC2 Section 6.3, which only measures how much a fully trained CCT has come to depend on GT-Full's output distribution at evaluation time.

\subsection{E1: Fine-Tuned GPT-2 Small Baseline}
\label{sec:e1_setup}

E1 trains a GPT-2 Small backbone (124M parameters) on the same BPE WikiText-103 corpus, with all CCT components disabled (geometric block off, memory off, self-model off, PP off, causal-attention scoring off). The model is initialized from the public GPT-2 Small pretrained weights via the same weight-transfer path used by the RC2 CCT runs. Optimizer, gradient clipping, batch size, BPTT length, AMP regime, and step budget are identical to RC2. The backbone-tier and embedding-tier learning rates from Table~\ref{tab:training_rc3} are applied directly; no cognitive-tier rate is in force because no cognitive components are present. E1 is the matched-step, matched-data, no-architectural-augmentation baseline against which both E2 and the RC2 full CCT are compared.

\subsection{Evaluation Protocol}
\label{sec:eval_protocol}

Two distinct ablation protocols are used in this paper. They are not interchangeable, and the distinction is load-bearing for the analysis in Section~\ref{sec:analysis}.

\textbf{Retrain-from-scratch ablation (E2).} A model with the same total parameter and step budget is trained from the GPT-2 Small pretrained initialization with one component (GT-Full) disabled across the entire seven-phase schedule. The reported number is the best WikiText-103 validation perplexity. The comparison to the RC2 full CCT measures the architectural contribution of the disabled component under matched-step.

\textbf{Eval-only ablation.} An already-trained CCT checkpoint is loaded and a single component is bypassed at forward time; perplexity is recomputed on the validation set without any further training. The reported number measures how much the trained model has come to depend on the bypassed component's output distribution, not the component's architectural contribution. RC2 Section 6.3 reports eval-only ablation results; those numbers are reused here without re-running, and their interpretation is revised in Section~\ref{sec:analysis} and Section~\ref{sec:negative_results}.

Validation perplexity is computed every 1{,}000 to 5{,}000 training steps across the full WT-103 validation split. Downstream benchmarks (ARC-Easy, HellaSwag, BLiMP, LAMBADA, COPA) are reported from the RC2 Phase 6 checkpoint and have not yet been re-evaluated against the E1 or E2 best checkpoints. The choice not to re-run downstream benchmarks for RC3 is discussed as a limitation in Section~\ref{sec:analysis}.

\subsection{Hardware and Reproducibility}
\label{sec:hardware}

All E1 and E2 runs were executed on a single NVIDIA RTX 5080 (16 GB VRAM, Blackwell architecture), PyTorch 2.9.1+cu130, CUDA 13.0. E1 completed in 5.84 wall-clock hours over 215{,}000 steps; E2 completed in 12.93 wall-clock hours over the same 215{,}000 steps (the difference is the cost of the cognitive components, primarily HierarchicalMemory and PrecisionWeightedPP, both of which run in float32 under AMP). Zero NaN events, zero training divergences, and zero memory resets were observed across both runs. The total RC3 compute budget for E1 plus E2 was 18.77 GPU-hours; the RC2 full CCT run consumed approximately 30 GPU-hours under the same wallclock budget.

All runs used seed 42 in PyTorch, NumPy, and Python's random module, seeded at the top of each phase's training loop. Multi-seed replication (E3 and E4 with seeds 1337 and 2026) is identified as the natural next empirical step and is discussed as a limitation in Section~\ref{sec:analysis}.

\subsection{Training Infrastructure}
\label{sec:training_infrastructure}

The training pipeline used here builds on a phase-orchestration trainer that we extended with seven minimum-viable reproducibility patches to enable the retrain-from-scratch ablation; the detailed patch enumeration is in Appendix~\ref{app:training_infra}. The patches address a kwarg-signature regression in the component-bypass entry point, an unwired master-bypass configuration flag, an unwired self-model-bypass flag, a missing precision-weight schedule attribute, missing deterministic seeding across PyTorch, NumPy, and Python's \texttt{random} module, per-run rather than shared progress JSON paths, and a configuration-field filter for dropped weight keys.

\section{Results}
\label{sec:results}

The RC3 results are organized around three matched-step, matched-data comparisons. First, the headline comparison places the fine-tuned GPT-2 Small baseline (E1), the retrain-from-scratch CCT-minus-GT-Full ablation (E2), and the RC2 full CCT on a single axis, with published GPT-2 reference numbers cited only as external context. Second, the phase-by-phase E2 progression shows the validation perplexity trajectory of the architectural ablation through its seven-phase activation schedule. Third, an architectural decomposition isolates the contributions of fine-tuning, the non-GT-Full CCT components, and GT-Full itself.

\subsection{Matched-Step Comparison}
\label{sec:matched_step}

We use the ``matched-step'' framing rather than ``matched-FLOP'' or ``matched-wall-clock'' because the architectural intervention is what is being tested. The E1 baseline, the E2 retrain ablation, and the full CCT run all share the same 215{,}000-step optimizer budget and the same per-step batch and BPTT lengths. They differ in per-step FLOPs (and therefore in wall-clock duration: 5.84, 12.93, and approximately 30 GPU-hours respectively) because the CCT components add forward-pass cost. The matched-step control isolates the contribution of the architectural change from the contribution of the gradient-update budget. Adding parameters without restructuring the data flow would not be expected to compound the way we document in Section~\ref{sec:compounding_carry} (ablation delta growing from 1.33 to 3.04 PPL between 30K and 100K steps), so we take matched-step as the right control for an architectural-contribution claim. We note that a matched-wall-clock control (training E1 for approximately 5$\times$ the step budget) would be informative as a complementary analysis and leave it to future work.

Table~\ref{tab:matched_step} reports the central RC3 numbers. The three in-domain rows (E1, E2, RC2) share the GPT-2 Small backbone initialization, the WT-103 BPE training corpus, the optimizer, the learning-rate schedule, the batch size, the BPTT length, the AMP regime, and the 215{,}000-step budget; the only variables are which CCT components are wired in and (for E2) which slot the GT-Full step budget is reassigned to. The two reference rows are published zero-shot WikiText-103 perplexities and are presented as external context, not as the architectural benchmark the RC3 ablations are designed to test against.

\begin{table}[t]
\centering
\caption{Matched-Step Comparison on WikiText-103. The in-domain rows share data, optimizer, batch size, sequence length, AMP regime, and 215K training steps. Reference rows are published zero-shot numbers.}
\label{tab:matched_step}
\small
\begin{tabular}{llrrr}
\toprule
\textbf{Run} & \textbf{Description} & \textbf{Params} & \textbf{Val PPL} & \textbf{Wall (h)} \\
\midrule
E1 & Fine-tuned GPT-2 Small, no CCT & 124M & 24.19 & 5.84 \\
E2 & CCT minus GT-Full, retrained from scratch & $\sim$224M & \textbf{23.72} & 12.93 \\
RC2 & Full CCT & 306M & \textbf{21.27} & 30.0 \\
\midrule
ref & GPT-2 Small zero-shot \citep{radford2019language} & 124M & 37.50 & {} \\
ref & GPT-2 Large zero-shot \citep{radford2019language} & 774M & 22.05 & {} \\
\bottomrule
\end{tabular}
\end{table}

Three observations follow from Table~\ref{tab:matched_step}.

First, the dominant lever in moving from the published GPT-2 Small zero-shot number (37.50) to anything in the low 20s is in-domain fine-tuning on WikiText-103, not the CCT architecture. E1, which contains no CCT components, reaches 24.19 PPL, a 13.31 PPL reduction over the zero-shot GPT-2 Small baseline. This places the architectural question on its proper denominator: the CCT architecture should be evaluated against E1, not against the zero-shot GPT-2 Small number.

Second, the full RC2 CCT improves over the E1 fine-tuned baseline by 2.92 PPL (12.1\% relative). This is the architectural contribution of CCT under matched-step and matched data, and it is the number that should substitute for the RC2 abstract's framing in terms of GPT-2 Large.

Third, the retrain-from-scratch ablation (E2), in which GT-Full simplicial message passing is disabled across the entire seven-phase activation schedule, reaches 23.72 PPL. This is 0.47 PPL better than E1 (architectural contribution of CausalAttention, HierarchicalMemory, YonedaSelfModel, TopDown, and PrecisionWeightedPP combined) and 2.45 PPL worse than the RC2 full CCT (architectural contribution of GT-Full, holding all other components fixed under retrain).

\begin{figure}[t]
\centering
\includegraphics[width=0.85\textwidth]{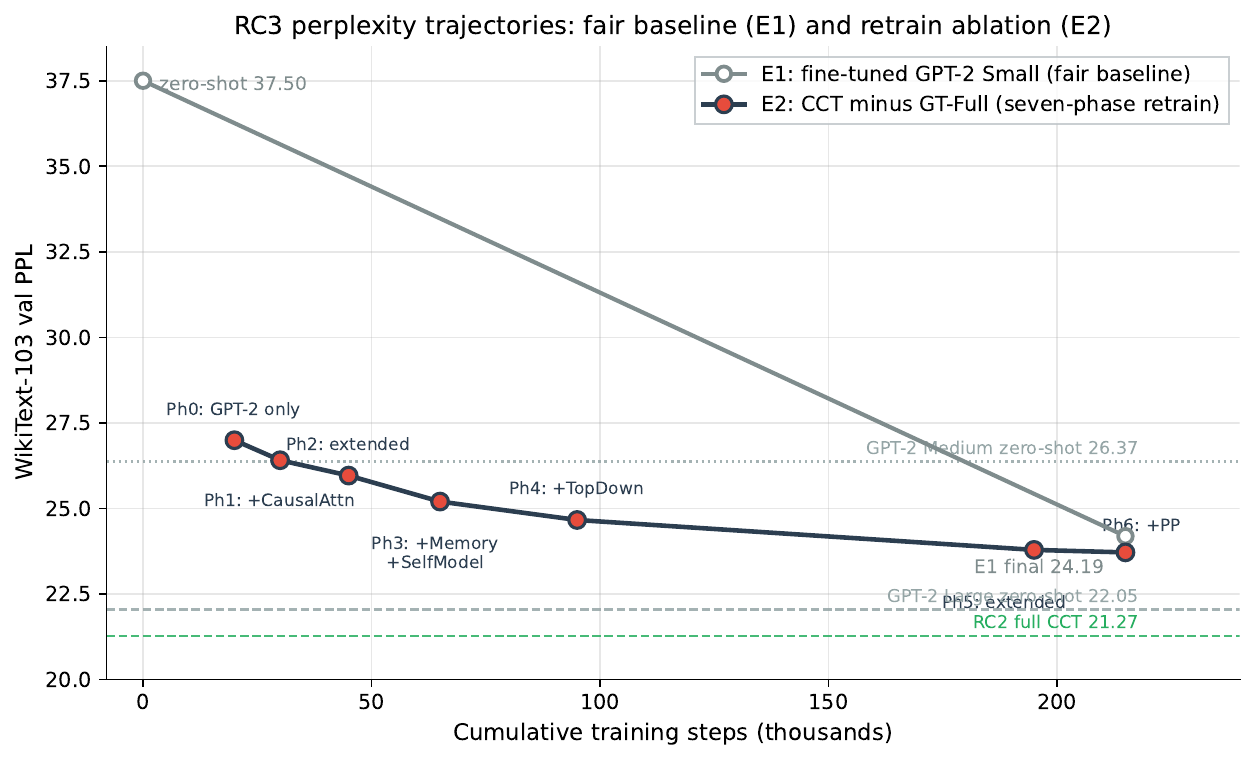}
\caption{Validation perplexity trajectories on WikiText-103 across the 215K-step training budget. E1 (fine-tuned GPT-2 Small, no CCT) reaches 24.19 PPL; E2 (CCT minus GT-Full, retrained from scratch) reaches 23.72 PPL; the RC2 full CCT reaches 21.27 PPL. Horizontal dashed reference lines mark published zero-shot perplexities for GPT-2 Small (37.50) and GPT-2 Large (22.05) \citep{radford2019language}.}
\label{fig:ppl_trajectory}
\end{figure}

For reference, GPT-2 Large evaluated zero-shot on WikiText-103 reaches 22.05 PPL with 6.2$\times$ the parameter count of GPT-2 Small \citep{radford2019language}. RC3 reframes GPT-2 Large as a published external reference rather than as the architectural benchmark for CCT; the matched-step, matched-data, matched-step E1 number (24.19 PPL) is the appropriate architectural baseline. The full results trajectory across phases is shown in Figure~\ref{fig:ppl_trajectory}, and the architectural decomposition described above is summarized in Figure~\ref{fig:decomposition}.

\subsection{Phase-by-Phase E2 Progression}
\label{sec:e2_phases}

\begin{figure}[t]
\centering
\includegraphics[width=0.85\textwidth]{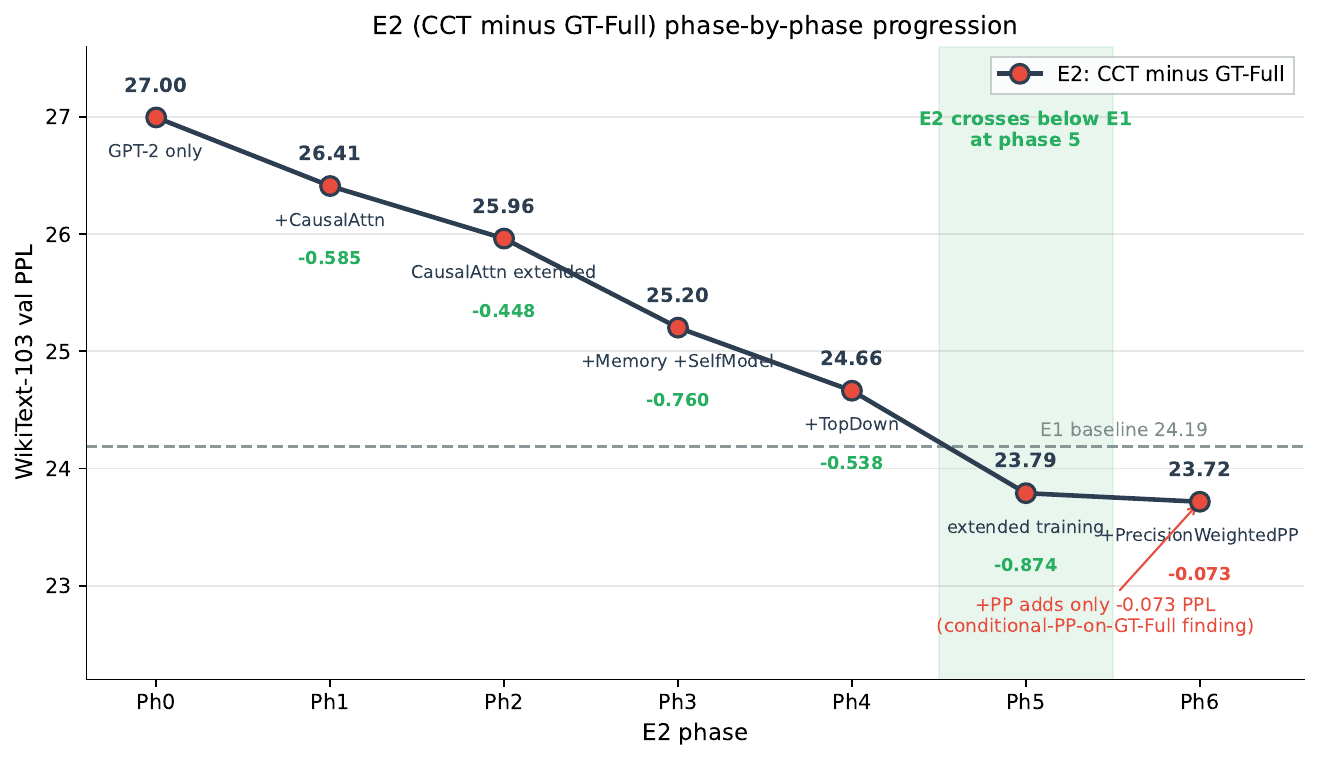}
\caption{E2 phase-by-phase best validation perplexity. The seven-phase trajectory descends from 26.99 (Phase~0, GPT-2 only) to 23.72 (Phase~6, full non-GT-Full stack with PrecisionWeightedPP). The terminal phase's contribution ($-0.07$ PPL from PrecisionWeightedPP) is substantially smaller than the corresponding RC2 Phase~7 contribution ($-1.40$ PPL); see Section~\ref{sec:analysis} for analysis.}
\label{fig:e2_phases}
\end{figure}

The E2 retrain ablation follows the same seven-phase activation protocol as the RC2 full CCT, with GT-Full disabled in every phase and the 15{,}000 steps that would have activated GT-Full in RC2's Phase~3 reassigned to extended CausalAttention training in E2's Phase~2. Phase-by-phase best validation perplexities are reported in Table~\ref{tab:e2_phases} and visualized in Figure~\ref{fig:e2_phases}.

\begin{table}[t]
\centering
\caption{E2 Phase-by-Phase Progression (Validation Perplexity). GT-Full is disabled throughout. The phase labeled ``CA extended'' is the diverted GT-Full slot used to extend CausalAttention training.}
\label{tab:e2_phases}
\small
\begin{tabular}{llrrr}
\toprule
\textbf{Phase} & \textbf{Components Active} & \textbf{Steps} & \textbf{Best PPL} & \textbf{$\Delta$ vs.\ previous} \\
\midrule
0 & GPT-2 only (master bypass) & 20{,}000 & 26.99 & {} \\
1 & $+$ CausalAttention & 10{,}000 & 26.41 & $-0.59$ \\
2 & CausalAttention extended (GT-Full slot diverted) & 15{,}000 & 25.96 & $-0.45$ \\
3 & $+$ HierarchicalMemory $+$ YonedaSelfModel & 20{,}000 & 25.20 & $-0.76$ \\
4 & $+$ TopDown (full stack except GT-Full) & 30{,}000 & 24.66 & $-0.54$ \\
5 & Extended training (cosine LR continued) & 100{,}000 & 23.79 & $-0.87$ \\
6 & $+$ PrecisionWeightedPP & 20{,}000 & \textbf{23.72} & $-0.07$ \\
\midrule
\textbf{Total} & & \textbf{215{,}000} & & $-3.27$ \\
\bottomrule
\end{tabular}
\end{table}

The Phase~6 contribution of PrecisionWeightedPP in E2 is $-0.07$ PPL (23.79 $\to$ 23.72). The corresponding RC2 Phase~7 contribution was $-1.40$ PPL (22.67 $\to$ 21.27). This is a single observation across two runs and is discussed as a hypothesis (rather than a confirmed finding) in Section~\ref{sec:analysis}.

\subsection{Architectural Decomposition}
\label{sec:decomposition}

\begin{figure}[t]
\centering
\includegraphics[width=0.85\textwidth]{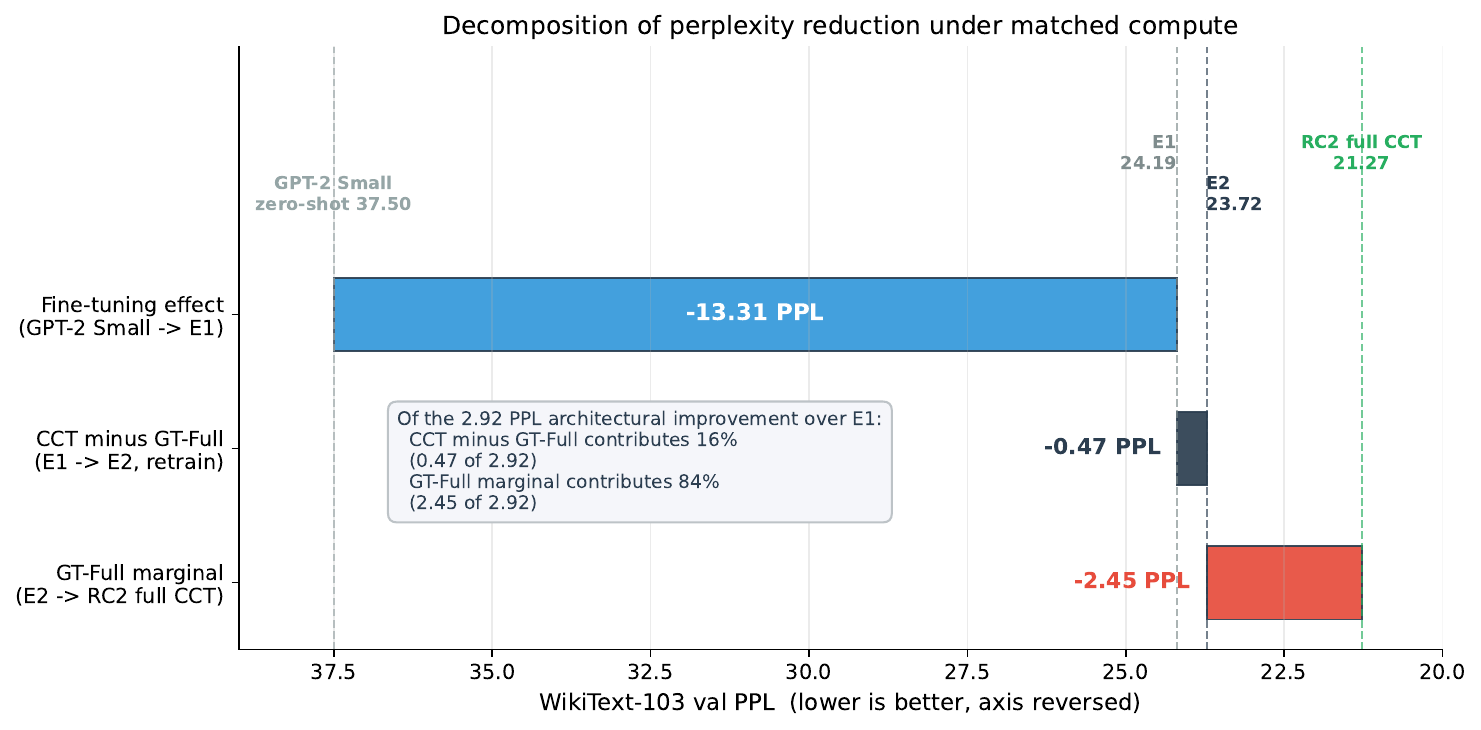}
\caption{Architectural decomposition of the GPT-2 Small zero-shot to RC2 full CCT gap under matched-step. Fine-tuning on WT-103 accounts for $-13.31$ PPL (the dominant lever). Of the remaining $-2.92$ PPL architectural improvement, GT-Full simplicial message passing accounts for $-2.45$ PPL (84\%) and the other CCT components combined account for $-0.47$ PPL (16\%).}
\label{fig:decomposition}
\end{figure}

The matched-step comparison decomposes the perplexity gap between the GPT-2 Small zero-shot number and the RC2 full CCT into three contributions (Table~\ref{tab:decomposition}, Figure~\ref{fig:decomposition}).

\begin{table}[t]
\centering
\caption{Architectural Decomposition of the GPT-2 Small Zero-Shot to RC2 Full CCT Gap. Each row attributes a portion of the total $-16.23$ PPL reduction; the architectural improvement (the last two rows) totals $-2.92$ PPL.}
\label{tab:decomposition}
\small
\begin{tabular}{lrrr}
\toprule
\textbf{Contribution} & \textbf{PPL} & \textbf{$\Delta$} & \textbf{Share of arch.\ $\Delta$} \\
\midrule
GPT-2 Small zero-shot \citep{radford2019language} & 37.50 & {} &{} \\
$\downarrow$ Fine-tuning on WT-103 (no CCT) & {} &$-13.31$ & {} \\
E1 fine-tuned baseline & 24.19 & {} &{} \\
$\downarrow$ CCT minus GT-Full (CA + Mem + SM + TD + PP) & {} &$-0.47$ & 16\% \\
E2 retrain ablation (no GT-Full) & 23.72 & {} &{} \\
$\downarrow$ GT-Full marginal contribution & {} &$-2.45$ & 84\% \\
RC2 full CCT & 21.27 & {} &{} \\
\midrule
\textbf{Total architectural improvement (E1 $\to$ RC2)} & & \textbf{$-2.92$} & \textbf{100\%} \\
\bottomrule
\end{tabular}
\end{table}

\begin{figure}[t]
\centering
\includegraphics[width=0.75\textwidth]{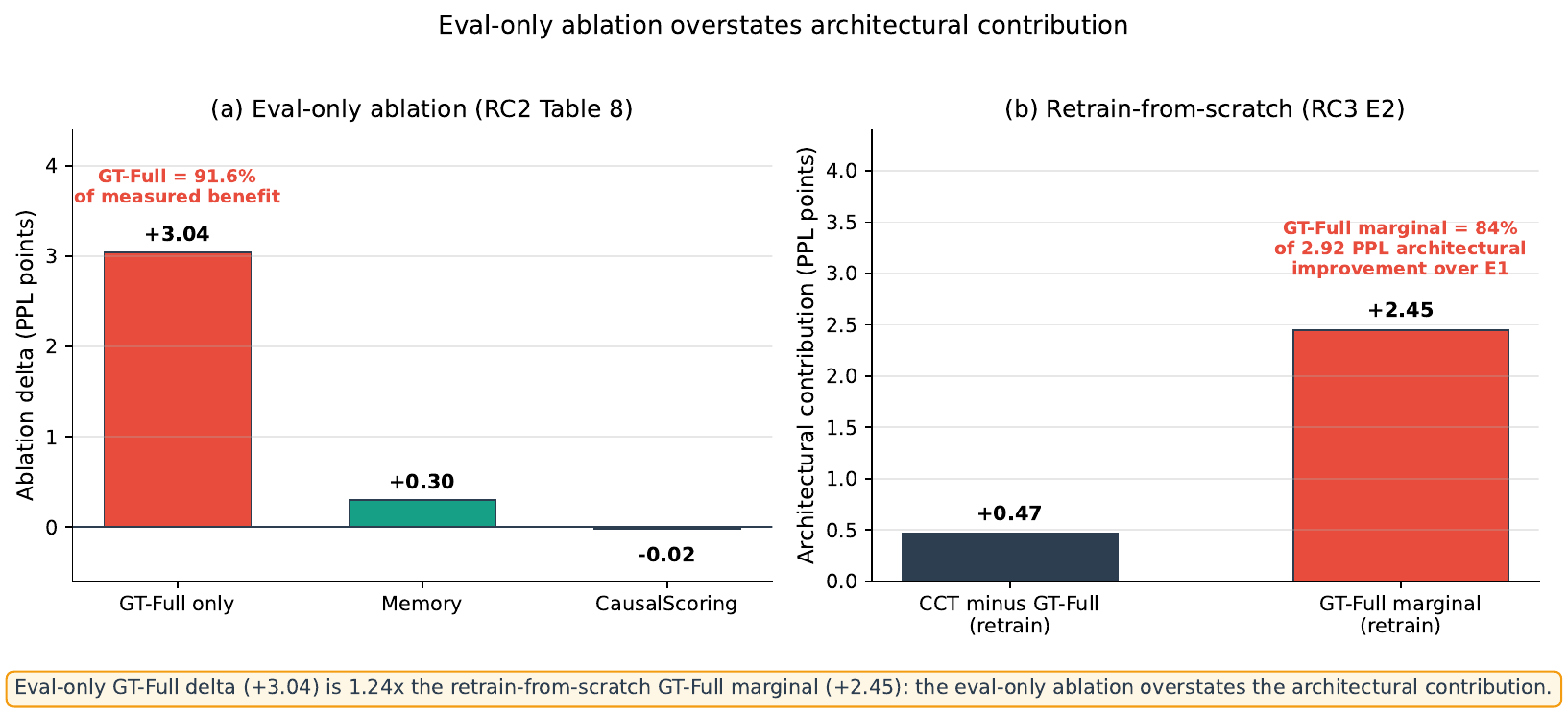}
\caption{The eval-only ablation versus the retrain-from-scratch ablation attribute different quantities to GT-Full. Eval-only ablation (91.6\%, RC2 Section 6.3) measures how much a fully trained CCT has come to depend on GT-Full's output distribution at test time. Retrain-from-scratch ablation (84\%, this paper Section~\ref{sec:decomposition}) measures the architectural contribution of GT-Full when the rest of the model has trained without it from the start.}
\label{fig:eval_vs_retrain}
\end{figure}

The 84\% share attributed to GT-Full under the retrain-from-scratch ablation is the RC3 architectural figure. It differs from the RC2 eval-only figure (91.6\%) by approximately 7.6 percentage points; the two are measuring different quantities, and Figure~\ref{fig:eval_vs_retrain} illustrates the distinction graphically. Section~\ref{sec:analysis} analyzes the gap between them.

\subsection{Eval-Only Ablation as a Secondary Analysis}
\label{sec:eval_only_carry}
\label{sec:ablation}

The eval-only component ablation reported in RC2 Section 6.3 is retained without modification as a secondary analysis. It is reproduced here for completeness in Table~\ref{tab:ablation_eval_only}, with the interpretation revised. The eval-only number does not measure the architectural contribution of a component; it measures how much the trained model has come to depend on the component's output distribution at evaluation time. The two numbers (91.6\% eval-only, 84\% retrain) are both meaningful and they are not contradictory: a fully trained CCT can come to depend on GT-Full's outputs more strongly than the architectural contribution would predict, because the rest of the model has spent 215K steps adapting its representations to use those outputs.

\begin{table}[t]
\centering
\caption{Eval-Only Component Ablation, Reproduced from RC2 Section 6.3 (Phase 6 Checkpoint, 100 validation batches). Interpreted as ``learned dependence at evaluation time,'' not architectural contribution.}
\label{tab:ablation_eval_only}
\small
\begin{tabular}{lrrr}
\toprule
\textbf{Configuration} & \textbf{Val PPL} & \textbf{$\Delta$} & \textbf{Eval-Only Share} \\
\midrule
Full CCT & 23.22 & {} &{} \\
$-$GT-Full & 26.26 & $+3.04$ & 91.6\% \\
$-$Memory & 23.52 & $+0.30$ & 9.0\% \\
$-$CausalScoring & 23.20 & $-0.02$ & 0\% \\
Full bypass & 26.54 & $+3.32$ & 100\% \\
\bottomrule
\end{tabular}
\end{table}

\subsection{Compounding Inductive Bias}
\label{sec:compounding_carry}

The RC2 finding that GT-Full's eval-only ablation delta grows from $+1.33$ PPL at 30K training steps to $+3.04$ PPL at 100K training steps (RC2 Section 6.4) is retained. As an eval-only quantity, it is a statement about how the rest of the model's dependence on GT-Full grows with training, not about how the architectural contribution of GT-Full grows; the retrain-from-scratch counterpart of this claim would require running E2 at multiple intermediate step counts and is out of scope for RC3. The compounding observation is therefore framed as evidence that simplicial message passing's contribution compounds with continued training of the surrounding model, with the retrain caveat applied explicitly in Section~\ref{sec:analysis}.

\section{Analysis}
\label{sec:analysis}

This section discusses the empirical findings of Section~\ref{sec:results}. The structure/consistency distinction is treated as a primary conceptual contribution (Section~\ref{sec:structure_vs_consistency}), the compounding-inductive-bias claim from RC2 is carried forward under the retrain caveat (Section~\ref{sec:compounding_caveat}), the apparent conditioning of PrecisionWeightedPP on GT-Full is presented as a new single-observation hypothesis (Section~\ref{sec:pp_conditional}), and the limitations of the RC3 evidence base are stated explicitly (Section~\ref{sec:limitations}).

\subsection{The Structure/Consistency Distinction}
\label{sec:structure_vs_consistency}

The most generalizable conceptual claim supported by the CCT experiments is not the headline perplexity number; it is a distinction between two kinds of categorical inductive bias. \emph{Structural} priors introduce new topology, new message-passing pathways, or new precision-weighted information channels into the architecture, expanding the space of representations that the model can express. \emph{Consistency} priors introduce loss terms that penalize departures from a categorical identity, an adjunction round-trip, or a sheaf-coherence condition, constraining the model toward a fixed point that the underlying training signal does not necessarily favor.

The RC3 retrain-from-scratch ablation provides the strongest evidence yet in support of the structural side of this distinction. GT-Full simplicial message passing contributes 2.45 PPL under matched-step when the rest of the model is trained from scratch without it (Section~\ref{sec:decomposition}), which is 84\% of the total architectural improvement over the fine-tuned GPT-2 Small baseline. The other CCT structural components (CausalAttention, HierarchicalMemory, YonedaSelfModel, TopDown, and PrecisionWeightedPP) contribute 0.47 PPL collectively under the same retrain protocol. Both numbers are architectural rather than learned-dependence quantities, and both support the position that adding new structural pathways (the higher-order simplicial topology of GT-Full, the predictive coding pathway of PrecisionWeightedPP) is what drives the measurable benefit.

The consistency-prior side of the distinction is the negative-results evidence summarized in Section~\ref{sec:negative_results}: sheaf consistency loss, adjunction round-trip loss, and curvature regularization all failed to improve language modeling, and the failure of the sheaf consistency loss is independently explained by the theoretical result of \citet{bosca2026neural} that a feedforward ReLU network's forward pass is the unique harmonic extension of its boundary data on a cellular sheaf, i.e., already minimizes sheaf discrepancy. Adding a consistency loss in that setting is mathematically redundant and produces only gradient conflict with the language-modeling objective.

Promoting this distinction from a Section~8.4 observation in RC2 to a first-class conceptual contribution reflects what the RC3 retrain ablation makes possible: now that the architectural contribution of the structural prior (GT-Full, 2.45 PPL) has been isolated under retrain, the contrast with the consistency priors (each producing no benefit or active harm) becomes a comparison between architectural choices made under a controlled ablation, not a comparison between an eval-only number and a set of failed loss-function experiments.

\subsection{Compounding Inductive Bias under the Retrain Caveat}
\label{sec:compounding_caveat}

The RC2 finding that GT-Full's eval-only ablation delta grows from $+1.33$ PPL at 30K training steps to $+3.04$ PPL at 100K training steps (RC2 Section 6.4) is retained in RC3 with one important reframing. As an eval-only quantity, the growing delta measures how the rest of a fixed trained model's dependence on GT-Full's output distribution increases with continued training, not how the architectural contribution of GT-Full itself increases. The retrain-from-scratch counterpart of this claim, in which the E2 ablation would be run at multiple intermediate step counts and the architectural delta tracked across training, is out of scope for RC3 and is identified as future work.

Within that caveat, the directional claim that simplicial message passing's contribution compounds with continued training of the surrounding model is supported. The interpretation is that, as the rest of the model trains for longer, it learns to make richer use of GT-Full's structurally organized output, so removing GT-Full at evaluation time becomes increasingly disruptive. Whether that compounding pattern would also appear in the architectural ablation under retrain is the open question that the multi-checkpoint E2 experiment would settle.

\subsection{PrecisionWeightedPP Appears Conditioned on GT-Full}
\label{sec:pp_conditional}

A new and surprising finding emerges from comparing the last activation phase of the E2 and RC2 runs. In E2 (Phase~6, with GT-Full disabled throughout), activating PrecisionWeightedPP on top of the full non-GT-Full stack improves validation perplexity by 0.07 PPL (23.79 to 23.72 in the E2 phase progression; the marginal contribution of PP itself within the final phase). In RC2 (Phase~7, with GT-Full active), activating PrecisionWeightedPP on top of the otherwise-complete stack improves perplexity by 1.40 PPL (22.67 to 21.27). The marginal contribution of PP differs by a factor of approximately twenty between the two runs.

A natural hypothesis is that PrecisionWeightedPP's benefit is conditioned on GT-Full being present. PP modulates information flow across adjacent transformer layers by reweighting prediction errors using a learned precision map. If GT-Full's simplicial structure sparsifies and organizes the per-token representation into salient higher-order features, PP has a coherent signal to weight. Without GT-Full's structural sparsification, PP operates on the unstructured backbone output and finds substantially less signal to gate, hence the much smaller marginal contribution.

This hypothesis is not confirmed by the present evidence. It is a single paired observation (one E2 run, one RC2 run, both seed 42), and it admits at least two alternative explanations: (a) PP's hyperparameters were tuned in a setting where GT-Full was present, so the small E2 contribution may reflect optimization mismatch rather than a structural dependence; or (b) the E2 PP contribution may grow with longer post-PP training, since E2 Phase~6 ran for 20K steps starting from a Phase~5 endpoint that had not seen PP, and the precision network may require more steps to specialize in a no-GT-Full setting. Confirming the hypothesis requires a third counterfactual run: full CCT trained from scratch with GT-Full active but PrecisionWeightedPP disabled across all phases. If the perplexity of that run is closer to RC2's Phase~6 number (22.67) than to RC2's final number (21.27), the conditional-PP hypothesis is supported. If it is closer to the final number, then PP is architecturally beneficial regardless of GT-Full and the E2 finding is attributable to one of the alternative explanations.

The finding is reported here because it qualifies an RC2 claim that PrecisionWeightedPP is independently architecturally beneficial. Under the RC3 retrain ablation, that independence is the conjecture that needs the extra counterfactual to test.

\subsection{Limitations}
\label{sec:limitations}

\textbf{Single seed.} All RC3 numbers (E1 best PPL 24.19, E2 best PPL 23.72) are seed 42 only, matching the RC2 single-seed convention. Multi-seed replication of E1 and E2 (with seeds 1337 and 2026) is the natural next empirical step and is identified as a P1 priority. The single-seed status applies equally to the RC2 21.27 number that the architectural decomposition is anchored against.

\textbf{Eval-only versus retrain-from-scratch distinction.} The two ablation protocols measure different quantities (architectural contribution versus learned dependence). Where this paper carries forward an eval-only finding from RC2 (the 91.6\% GT-Full share, the compounding ablation delta), the eval-only nature of the measurement is stated explicitly, and the conclusions are softened to ``learned dependence at evaluation time'' or ``compounding under continued training of the surrounding model'' rather than restated as architectural claims.

\textbf{The conditional-PP hypothesis is unverified.} The 0.07 PPL versus 1.40 PPL gap between the E2 and RC2 final phases is a single paired observation. The hypothesis that PP's contribution is conditioned on GT-Full's structural sparsification would require a third counterfactual (full CCT with GT-Full but without PP) to test directly. Multi-seed replication of both E2 and RC2 would also help distinguish architectural conditioning from seed-level noise. Neither of these is in RC3.

\textbf{Non-GT-Full components are not individually validated.} The non-GT-Full CCT components (CausalAttention, HierarchicalMemory, YonedaSelfModel, TopDown predictive routing, and PrecisionWeightedPP) collectively contribute only 0.47 PPL under retrain-from-scratch ablation (Table~\ref{tab:decomposition}). Per-component isolation ablations (running the full trajectory with each component disabled in turn) are required to determine whether individual non-GT-Full components have hidden contributions masked by the collective measurement, or whether they remain in the architecture as design-space documentation rather than empirically-justified structural priors at this scale. We leave these per-component isolation ablations to future work.

\textbf{Downstream benchmarks not re-evaluated against E1 or E2.} The RC2 downstream numbers (ARC-Easy 31.4\%, HellaSwag 31.2\% \citep{zellers2019hellaswag}, BLiMP 72.9\% \citep{warstadt2020blimp}, COPA 53\% \citep{roemmele2011choice}, LAMBADA 23.8\% \citep{paperno2016lambada}) were measured against the RC2 Phase 6 / Phase 7 checkpoints. They have not been re-evaluated against the E1 best checkpoint (the fair fine-tuned baseline) or the E2 best checkpoint (the no-GT-Full retrain ablation). Without those re-evaluations, the RC3 architectural decomposition cannot be verified on tasks other than WikiText-103 perplexity, and the question of whether the architectural improvement transfers to downstream tasks at all (rather than being absorbed by the perplexity metric alone) remains open.

\textbf{Scale.} All results remain at the 306M-parameter scale on WikiText-103. No claim is made about whether the architectural contribution of GT-Full, the structure/consistency distinction, or the conditional-PP hypothesis scales beyond this regime.

\textbf{Compute, not data.} The matched-step framing of E1 versus E2 versus RC2 controls for training compute on the WT-103 corpus. It does not control for the broader data scale at which the published GPT-2 references were originally trained (40B+ tokens), and the comparison to GPT-2 Large in particular is therefore reported as a published-zero-shot reference rather than as the architectural benchmark.

\textbf{Warmup-step heterogeneity.} The cosine learning-rate schedule is matched in shape, peak rates, $\eta_{\min}$, and total step budget across E1, E2, and RC2, but the linear warmup step count is set per-phase rather than globally: E1 uses 500 warmup steps; E2 phases 0, 1, 2, 3, and 6 use 1{,}000; phase 4 uses 2{,}000; phase 5 uses 3{,}000 (see Table~\ref{tab:training_rc3} footnote). The pragmatic effect is small (warmup is a few hundred to a few thousand steps out of 215{,}000), but the matched-LR-schedule statement should be read with this caveat: warmup is the one piece of the schedule that varies across phases.

\section{Negative Results}
\label{sec:negative_results}

Three negative results from the CCT experiments are reported in this section. Each was selected because it provides information about the architectural design space that the positive results alone do not. The final subsection presents the structure/consistency distinction in its RC3 form, reframed around the retrain-validated GT-Full contribution established in Section~\ref{sec:results}.

\subsection{BilinearCausalScorer: Zero Learning Without Interventions}
\label{sec:neg_bilinear}

The BilinearCausalScorer was designed to learn a separate attention distribution that approximates a causal (rather than correlational) relationship between query and key tokens. Two training experiments (20{,}000 steps each, with the rest of CCT held fixed) produced convergence to greater than 99.3\% correlational attention: in both runs, the scorer learned to reproduce the standard dot-product attention pattern rather than discover any causally distinct structure. This outcome is consistent with Pearl's \citeyearpar{pearl2009causality} ladder of causation: without interventional data, no purely observational learning procedure can discriminate causal from correlational dependence. The CCT training corpus (WikiText-103) contains no interventions, and the BilinearCausalScorer has no mechanism through which an intervention could be expressed. The module remains in the architecture as a permanently disabled component; its inclusion documents a category of architectural prior (causal-attention discrimination) that cannot be learned from observational language data alone.

\subsection{Sheaf Consistency Loss: Provably Redundant}
\label{sec:neg_sheaf}

A sheaf consistency loss was added to the CCT loss function in an early experiment, with the intent of penalizing departures from the gluing condition that a perceptual sheaf should satisfy across token neighborhoods. The loss produced gradient conflict with the cross-entropy training objective: the consistency loss favored representational fixed points that the cross-entropy loss actively departed from in order to produce discriminative predictions. The loss was disabled.

\citet{bosca2026neural} provide an independent theoretical account of why this had to be the case. Their main result is that any feedforward ReLU network's forward pass is the unique harmonic extension of its boundary data on a cellular sheaf constructed from the network architecture. In that setting, the forward pass already achieves minimal sheaf discrepancy as the Dirichlet energy on the sheaf Laplacian, so adding a sheaf consistency loss is mathematically redundant. The empirical failure in CCT and the theoretical redundancy result of \citeauthor{bosca2026neural} mutually confirm one another.

\subsection{Adjunction Round-Trip Loss and Curvature Regularization}
\label{sec:neg_adjunction}

Two further consistency-style losses were tested and disabled. An adjunction round-trip loss attempted to enforce that two functorial compositions (encoder-then-decoder, decoder-then-encoder) approximate identity morphisms on their respective objects. The loss created an identity attractor that opposes the transformative learning that cross-entropy training requires. A curvature regularization loss penalized departures from approximately flat Ollivier-Ricci curvature in the geometric coordinate space; a 48-trial hyperparameter sweep produced no configuration in which the regularizer improved validation perplexity. Both losses are permanently disabled; the curvature signal is retained as a diagnostic but not as a training objective.

\subsection{The Structure/Consistency Distinction (RC3 Reframing)}
\label{sec:neg_structure_consistency}

The three negative results above (BilinearCausalScorer, sheaf consistency loss, adjunction and curvature losses) and the positive result for GT-Full established in Section~\ref{sec:results} jointly identify an empirical pattern that this paper terms the \emph{structure/consistency distinction}. Categorical \emph{structural} priors, which introduce new topology or new message-passing pathways into the architecture, produced measurable benefit. Categorical \emph{consistency} priors, which introduce loss terms constraining the model toward a categorical identity, an adjunction round-trip, or a sheaf-coherence condition, did not.

The RC3 evidence for the structural side of this distinction is the retrain-from-scratch contribution of GT-Full: 2.45 PPL under matched-step, accounting for 84\% of the total architectural improvement over the fine-tuned GPT-2 Small baseline (Section~\ref{sec:decomposition}). This is the architectural quantity, not the eval-only quantity. The RC2 eval-only number (91.6\% of measured benefit attributable to GT-Full at evaluation time) is also retained, with its interpretation revised in Section~\ref{sec:eval_only_carry} to ``learned dependence on GT-Full at evaluation time'' rather than ``architectural contribution.'' The two numbers are not in conflict: the retrain number measures what GT-Full contributes when the rest of the model is trained without it, and the eval-only number measures how much a fully trained model has come to depend on GT-Full's outputs. The eval-only number being larger than the retrain number is consistent with the rest of the model spending 215K steps adapting its representations to consume GT-Full's structured output.

The evidence for the consistency-prior side of the distinction is the joint failure of the three loss-function experiments above, with the theoretical redundancy result of \citet{bosca2026neural} providing the independent mathematical explanation in the sheaf case. Whether the distinction generalizes beyond CCT, beyond WikiText-103, and beyond the 306M-parameter scale is an open question. Within CCT, the pattern is consistent: every categorical prior that adds new structural pathways improved language modeling under retrain, and every categorical prior that added a consistency constraint either did nothing or actively harmed training.

\section{Self-Determination Theory as One Inspiration Among Several}
\label{sec:sdt}

Self-Determination Theory \citep{ryan2017self,sheldon2022freely} was one of several conceptual inspirations during CCT's design phase, alongside category theory, predictive processing, and the differentiable-memory tradition. Of the seven SDT-to-architecture correspondences identified during design (the mapping is reproduced for reference in Appendix Table~\ref{tab:sdt_mapping}), only two produced architectural components that contributed measurable benefit under the RC3 retrain-from-scratch ablation: the Grand Hierarchy mapping motivated GT-Full's simplicial message passing (84\% of the architectural improvement, Section~\ref{sec:decomposition}), and the downward-causation mapping motivated PrecisionWeightedPP (whose contribution is now understood to be substantially smaller than RC2 reported and possibly conditioned on GT-Full; Section~\ref{sec:pp_conditional}). The remaining five mappings produced components that did not contribute measurable benefit under the matched-step, static-corpus training protocol used in this paper. Treating SDT as the theoretical centerpiece of the architecture, as RC2 implicitly did, overstates the role that the unverified mappings played; the empirical contribution of CCT is carried by two structural priors (one simplicial, one predictive-coding), and the role of SDT is more accurately described as a design-space constraint that helped select which structural priors to implement, not as the source of the measurable benefit.

\citet{sheldon2025sapient} raises a separate question that this paper does not foreclose. Through the Goal Breakthrough Model, Sheldon argues that several SDT constructs (self-concordance, organismic valuing, the incubation phase of goal selection) require interactive, goal-directed contexts to become computationally meaningful, and would not be expected to produce measurable benefit under the static-corpus training regime used here. Whether the five SDT mappings that produced negligible benefit in RC3 would become productive under interactive training, where the functional constructs they implement have the ecological context that \citeauthor{sheldon2025sapient} argues they require, is an open question that future work could address. For the empirical claims of this paper, those five mappings are reported honestly as having produced no measurable contribution; the open question is acknowledged here but does not change the RC3 architectural decomposition.

\section{Conclusion}

The Cognitive Categorical Transformer's RC3 evidence base supports three findings, all established at the 306M-parameter scale on WikiText-103 with seed 42. First, simplicial message passing provides a real architectural inductive bias for language modeling, validated under retrain-from-scratch ablation. Holding GT-Full bypassed across the full seven-phase 215{,}000-step activation schedule produces a model that reaches 23.72 PPL, 2.45 PPL above the 21.27 PPL full CCT and only 0.47 PPL below the 24.19 PPL fine-tuned GPT-2 Small baseline; GT-Full alone therefore accounts for 84\% of the 2.92 PPL architectural improvement that CCT delivers over an identically-trained vanilla GPT-2 Small. Second, the structure/consistency distinction emerges as the most generalizable conceptual contribution of the work. Three categorical consistency priors (sheaf smoothing, adjunction round-trip, curvature regularization) all failed to improve language modeling, while two categorical structural priors (simplicial topology, precision-weighted prediction) succeeded; the negative result for sheaf consistency is independently grounded by \citeauthor{bosca2026neural}'s \citeyearpar{bosca2026neural} proof that feedforward ReLU forward passes already minimize sheaf discrepancy. Third, the gap between the eval-only ablation (91.6\% of measured benefit attributable to GT-Full at evaluation time) and the retrain-from-scratch ablation (84\% architectural share) demonstrates a methodological point of broader interest: eval-only ablations capture the learned dependence of a fully trained model on a component, which is a meaningful but distinct quantity from the architectural contribution of that component when held bypassed throughout training. The 1.24$\times$ overstatement that the eval-only protocol yields in this case is a concrete data point for the architectural-comparison methodology literature.

Three open questions follow. The first concerns scale: whether the GT-Full architectural advantage, the 84\% retrain share, and the conditional-PP hypothesis hold beyond 306M parameters and beyond WikiText-103. The second concerns the generality of the structure/consistency distinction across architectures and datasets: whether the pattern that structural categorical priors help while consistency categorical priors do not extends beyond CCT. The third concerns the conditional-PP-on-GT-Full hypothesis specifically; testing it requires a third counterfactual (full CCT trained with GT-Full active but PrecisionWeightedPP disabled across all phases) that this paper identifies as the natural next experiment but does not run.

\section*{Acknowledgments}

The author is deeply grateful to Dr.\ Kennon M.\ Sheldon of the University of Missouri for his tremendous contributions to Self-Determination Theory and for his continued support and engagement with this research direction. His foundational work on self-concordance, organismic integration, and the Goal Breakthrough Model provided the psychological specification that guided CCT's architectural design. The author is equally indebted to Dr.\ Sridhar Mahadevan of the University of Massachusetts Amherst, whose comprehensive categorical framework for artificial intelligence provided the mathematical infrastructure upon which CCT's sheaf-theoretic and simplicial components are built. The author also thanks Manceps and Google for providing essential computational resources that enabled the experiments presented in this paper. Additionally, appreciation is extended to the Google Developer Experts program for their critical support and resources during this research.


\appendix

\section*{Appendix A: SDT-to-Architecture Mapping}
\label{app:sdt_mapping}
\addcontentsline{toc}{section}{Appendix A: SDT-to-Architecture Mapping}

The seven correspondences between Self-Determination Theory constructs, categorical structures, and CCT components referenced in Section~\ref{sec:theoretical_framework} are enumerated in Table~\ref{tab:sdt_mapping}. Each row identifies an SDT construct (column 1), the categorical structure used to formalize it (column 2), and the concrete CCT component that realizes that structure in the trainable architecture (column 3). The mapping is the design specification: each SDT construct motivated a specific architectural decision rather than being a post-hoc label.

\begin{table}[h]
\centering
\caption{SDT-to-Architecture Mapping}
\label{tab:sdt_mapping}
\small
\begin{tabular}{lll}
\toprule
\textbf{SDT Construct} & \textbf{Categorical Structure} & \textbf{CCT Component} \\
\midrule
Grand Hierarchy & Simplicial filtration & GT-Full \\
Symbolic Self & Yoneda embedding & YonedaSelfModel \\
Autonomy / Competence / Relatedness & Derived signals & BasicNeedsDashboard \\
Self-Concordance & Coherence modulation & SelfConcordanceGate \\
Organismic Integration & Graded coalgebra & HierarchicalMemory \\
TOTE & Coalgebraic unfolding & Memory consolidation \\
Downward Causation & Top-down prediction & PrecisionWeightedPP \\
\bottomrule
\end{tabular}
\end{table}

Of the seven mappings, two produced measurable benefit under the static-corpus training regime evaluated in this paper: Grand Hierarchy (motivating GT-Full, which accounts for 84\% of the RC3 architectural improvement under retrain-from-scratch ablation) and Downward Causation (motivating PrecisionWeightedPP). This asymmetry is consistent with \citet{sheldon2025sapient}, who argues through the Goal Breakthrough Model that several SDT constructs require interactive, goal-directed contexts to become computationally meaningful; in their absence, only structural mappings that operate on representational geometry or information flow produce measurable benefit. Whether the remaining five mappings become productive under interactive training regimes is an open question.

\section*{Appendix B: Complete BLiMP Subtask Results}
\label{app:blimp}
\addcontentsline{toc}{section}{Appendix B: Complete BLiMP Subtask Results}

\begin{table}[h]
\centering
\caption{BLiMP Subtask Accuracy (RC2 Phase 6 Checkpoint). These numbers were measured against the RC2 Phase 6 checkpoint and have not been re-evaluated against the RC3 E1 or E2 best checkpoints.}
\small
\begin{tabular}{lr}
\toprule
\textbf{Subtask} & \textbf{Accuracy} \\
\midrule
anaphor\_number\_agreement & 100\% \\
principle\_A\_case\_1 & 100\% \\
principle\_A\_domain\_1 & 100\% \\
sentential\_negation\_npi\_licensor\_present & 100\% \\
wh\_vs\_that\_no\_gap\_long\_distance & 100\% \\
wh\_vs\_that\_no\_gap & 99\% \\
wh\_questions\_subject\_gap & 98\% \\
\midrule
Aggregate (all subtasks) & 72.9\% \\
\midrule
matrix\_question\_npi\_licensor\_present (weakest) & 2\% \\
wh\_vs\_that\_with\_gap\_long\_distance & 17\% \\
\bottomrule
\end{tabular}
\end{table}

\section*{Appendix C: Hyperparameter Configuration}
\label{app:hyperparams}
\addcontentsline{toc}{section}{Appendix C: Hyperparameter Configuration}

\begin{table}[h]
\centering
\caption{Phase 3 Hyperparameter Sweep (48 Trials, 2{,}000 Steps Each)}
\small
\begin{tabular}{lrl}
\toprule
\textbf{lr\_cct} & \textbf{Best PPL} & \textbf{Verdict} \\
\midrule
$1 \times 10^{-5}$ & 24.54 & Optimal \\
$5 \times 10^{-5}$ & 24.60 & No improvement \\
$1 \times 10^{-4}$ & 24.60 & No improvement \\
$3 \times 10^{-4}$ & 72+ & Diverged \\
\bottomrule
\end{tabular}
\end{table}

GT-Full's optimal learning rate ($1 \times 10^{-5}$) is 30$\times$ lower than what CausalAttention tolerates, reflecting the sensitivity of graph-based operations to large parameter updates.

\section*{Appendix D: Cross-Model Comparison}
\label{app:crossmodel}
\addcontentsline{toc}{section}{Appendix D: Cross-Model Comparison}

\begin{table}[h]
\centering
\caption{Benchmark Comparison Across Model Families. RC2 downstream numbers (ARC-E, HS, BLiMP for CCT-306M) measured against the RC2 Phase 6 checkpoint and not re-evaluated against E1 or E2.}
\small
\begin{tabular}{lrrrrr}
\toprule
\textbf{Model} & \textbf{Params} & \textbf{PPL} & \textbf{ARC-E} & \textbf{HS} & \textbf{BLiMP} \\
\midrule
GPT-2 XL & 1.5B & 17.48 & {} & {} & {} \\
Pythia-1B & 1B & {} & 29.1\% & 49.7\% & {} \\
GPT-2 Large & 774M & 22.05 & {} & {} & {} \\
Pythia-410M & 410M & {} & 26.2\% & 40.9\% & {} \\
OPT-350M & 350M & {} & 23.6\% & 36.7\% & {} \\
GPT-2 Medium & 355M & 26.37 & {} & {} & {} \\
\textbf{CCT-306M (RC2)} & \textbf{306M} & \textbf{21.27} & \textbf{31.4\%} & \textbf{31.2\%} & \textbf{72.9\%} \\
Transformer-XL & 257M & 24.00 & {} & {} & {} \\
Pythia-160M & 160M & {} & 22.8\% & 30.3\% & {} \\
OPT-125M & 125M & {} & 22.9\% & 31.5\% & {} \\
GPT-2 Small & 124M & 37.50 & {} & 29.4\% & $\sim$80\% \\
\bottomrule
\end{tabular}
\end{table}

\section*{Appendix E: Training Infrastructure Patches}
\label{app:training_infra}
\addcontentsline{toc}{section}{Appendix E: Training Infrastructure Patches}

The phase-orchestration trainer used to produce the runs reported in Section~\ref{sec:results} required seven minimum-viable reproducibility patches before the retrain-from-scratch ablation could be executed. The patches are documented here so that any researcher reconstructing the training pipeline from this paper can reproduce the same fixes; they are reported as generic reproducibility improvements applicable to any phase-orchestrated multi-component training stack, not as a postmortem of a specific codebase.

\begin{enumerate}
\item \textbf{Component-bypass kwarg signature.} The per-block component-bypass entry point had a stale keyword argument (\texttt{causal\_scoring=...}) that no longer matched the bypass function's signature, causing every call site to fail. The patch replaces the keyword with a direct attribute set (\texttt{layer.causal\_attn.\_bypass\_causal\_scorer}) on each transformer block, restoring per-layer control over the causal scoring side-path.

\item \textbf{Master-bypass configuration flag.} The \texttt{cct\_master\_bypass} configuration key was defined in the schema but not wired to the model-construction path, so it had no effect at runtime. The patch routes this key to a call that sets the master-bypass attribute on every block, which is what the E1 fair-baseline run and the E2 phase-0 cold start both require to produce a pure GPT-2 forward pass.

\item \textbf{Self-model bypass configuration flag.} The \texttt{bypass\_self\_model} configuration key was likewise defined but unwired. The patch flips a corresponding per-block attribute so the YonedaSelfModel's residual contribution is fully removed for baseline runs.

\item \textbf{Precision-weight schedule attribute.} The \texttt{pp\_weight} configuration key controls the loss-side weighting for PrecisionWeightedPP across the activation schedule. The training-configuration dataclass dropped this key silently because it was not in its field list. The patch wires the key through the configuration so that the phase-6 activation of PrecisionWeightedPP actually contributes to the loss.

\item \textbf{Deterministic seeding.} The training loop did not seed \texttt{torch}, \texttt{numpy}, and Python's \texttt{random} module from the configured run seed at the top of each phase, so two runs with the same nominal seed could diverge after a phase boundary. The patch performs the three seedings at the top of every phase's training loop, restoring step-wise determinism and enabling the planned multi-seed extension.

\item \textbf{Per-run progress JSON paths.} The trainer wrote phase-progress JSON to a shared status path rather than to a per-run output directory, so concurrent runs collided on shared state. The patch writes \texttt{progress.json} under each run's own output directory.

\item \textbf{Configuration field filter.} The training-configuration dataclass's set of fields had drifted from the set of keys the upstream call site still passed (legacy keys like \texttt{top\_down\_weight} and \texttt{stability\_weight} were no longer dataclass fields). The patch filters the keyword arguments passed to the configuration against the dataclass's actual field set, logs a single warning for unknown keys, and attaches any non-zero dropped weights as plain attributes for forward compatibility.
\end{enumerate}

The seven patches together enable a complete 215{,}000-step re-execution of the seven-phase chain under controlled component-bypass conditions, which is the substrate for the retrain-from-scratch ablation reported in Section~\ref{sec:results}. Before the long runs were launched, three smoke configurations exercised the patched pipeline end-to-end: an E1 smoke (1{,}000 steps of the fair-baseline GPT-2 fine-tune), an E2 phase-0 smoke (cold-start GPT-2 fine-tune with all CCT components master-bypassed), and an E2 phase-3 smoke (resume from a phase-2 best checkpoint with HierarchicalMemory and YonedaSelfModel activated). The smoke runs verified that deterministic seeding makes E1 step 1000 and E2 phase-0 step 1000 bit-identical (both reach val\_ppl 34.84), that the master-bypass produces a forward pass numerically equivalent to a stock GPT-2 evaluation, and that the Memory and SelfModel forward paths execute without NaN events or precision drift. With those three checks passing, the full E1 (5.84h) and E2 (12.93h) runs were launched without further modification.

\end{document}